\documentclass[cameraready]{Interspeech}

\usepackage{amsmath, amssymb, graphicx, booktabs, enumitem, tikz, pgfplots, placeins, array, bm}
\usetikzlibrary{arrows.meta,positioning,fit}
\pgfplotsset{compat=1.18}

\newcommand{\CleanQ}{Clean-\ensuremath{Q}}
\newcommand{\BrokenQ}{Broken-\ensuremath{Q}}

\newcommand{\CleanQm}{\mathrm{Clean}\text{-}Q}
\newcommand{\BrokenQm}{\mathrm{Broken}\text{-}Q}
\newcommand{\NoQm}{\mathrm{No}\text{-}Q}

\title{When Does Quality-Aware Multimodal Fusion Matter? A Leakage-Safe Diagnostic for Decision-Level Dependence}

\author[affiliation={}, orcid=0009-0001-0251-6102]
{Jaden}{Moon}
\author[affiliation={}, orcid=0000-0002-2489-1130]
{Arvind}{Pillai}
\author[affiliation={}, orcid=0000-0001-7394-7682]
{Andrew}{Campbell}

\address{
    Dartmouth College, United States
}

\email{
    jaden.moon.27, arvind.pillai.gr, andrew.t.campbell\{@dartmouth.edu\}
}


\begin{document}

\maketitle

\begin{abstract}
Many multimodal systems estimate the reliability of each modality and weight their contributions to the final prediction. However, it remains unclear whether these scores influence model decisions or merely correlate with performance. We propose a simple diagnostic to test whether reliability information is used during inference. After training, the model and inputs are fixed while reliability scores are permuted across test examples. If predictions depend on these scores, performance should degrade. Experiments on StressID for stress recognition and CMU-MOSEI for sentiment analysis show that permuting reliability scores leaves performance unchanged despite substantial potential gains from selecting the best modality per example. In positive controls where reliability signals identify the correct modality, the same frozen fusion rules yield significant improvements, indicating that reliability signals influence fused decisions only when they reliably predict unimodal correctness.
\end{abstract}

\section{Introduction}

Multimodal systems frequently encounter heterogeneous evidence quality due to sensor failures, noise, and environmental interference \cite{Baltrusaitis.2018,nagaraj2023heterogeneity}. Quality-aware and uncertainty-aware fusion methods aim to handle this problem by weighting each modality according to an estimated reliability score. However, it remains unclear whether these scores actually influence model decisions at inference time, or whether observed improvements simply reflect correlations in the data \cite{tian2020uno,han2021trusted}. This raises a central question: when does conditioning on reliability actually change a multimodal prediction? We refer to these reliability estimates as \emph{quality signals} (\(Q\)).

This question is especially relevant for affect and stress recognition, where systems often combine speech, facial video, and physiological signals using early fusion, late fusion, or mixture-of-experts architectures \cite{han2021,zadeh2017,zhao22_interspeech,prisayad23_interspeech,tsai.2019,li.2024.correlated,liu.2025.missing}. Quality-aware weighting schemes have also been proposed for multimodal settings \cite{Baltrusaitis.2018,soleymani.2022,Li22.INT,schrufer24_interspeech}. Yet standard evaluations usually report whether a quality-aware architecture performs well, not whether the quality signal itself is decision-relevant. A model may improve because of architectural flexibility, dataset correlations, or missingness patterns, even if quality estimates are not used to decide which modality to trust \cite{Tellamekala2022,schuller21_compare}.

To test decision-level reliance directly, we formulate multimodal inference as \(p(y \mid E, M, Q)\), separating modality evidence \(E\), availability masks \(M\), and quality signals \(Q\). We introduce a leakage-safe, alignment-breaking diagnostic: after training, the model is frozen and quality values are shuffled across held-out examples while the sensory evidence, modality availability, and fusion rule stay fixed. The matched condition, \CleanQ{}, uses the observed quality values at inference. The shuffled condition, \BrokenQ{}, preserves the marginal distribution of quality values but breaks their instance-wise alignment with the corresponding inputs. To isolate quality effects from missingness, we evaluate on fully observed examples, where all modalities are present.

We evaluate this diagnostic primarily on StressID \cite{chaptoukaev2023stressid}, which contains heterogeneous modality availability, and additionally on CMU-MOSEI \cite{zadeh2018MOSEI} as a near-fully observed boundary case. Across these settings, shuffling native quality signals produces near-zero performance changes, even though an oracle that selects the best modality per example shows that better routing is possible. In contrast, positive controls produce large performance changes when quality is constructed to track corruption or unimodal correctness. These results suggest that quality-aware fusion affects decisions only when quality estimates identify which modality is reliable for the current instance.

\noindent\textbf{Contributions.}
\begin{itemize}
\item \textbf{Diagnostic.} We propose a leakage-safe post-hoc test that compares matched quality values with shuffled quality values under fixed experts and fusion parameters.
\item \textbf{Empirical finding.} On StressID, and in a near-fully observed CMU-MOSEI boundary case, shuffling native quality values produces negligible performance changes despite room for better per-example routing, indicating minimal decision-level reliance on the native quality signals.
\item \textbf{Validation.} Positive controls produce large permutation gaps when quality is aligned with corruption or unimodal correctness, demonstrating that the diagnostic detects reliance on quality when quality actually identifies the reliable expert.
\end{itemize}

\section{Decision-level identifiability diagnostic}
\label{sec:identifiability}

For each cross-validation fold, we split the data into training rows and held-out test rows. The diagnostic asks a targeted question: after a multimodal model has been trained, do its predictions change when only the match between quality estimates and test instances is broken?

\noindent\textbf{Separating evidence, availability, and quality.}
For modality $m$ and instance $i$, we represent the input using three components: evidence $E_{m,i}$, availability $M_{m,i}\in\{0,1\}$, and quality $Q_{m,i}$. Availability indicates whether the modality is observed, whereas quality estimates the reliability of the observed evidence. This separation distinguishes an \emph{outage}, where a modality is absent ($M_{m,i}=0$), from a \emph{degradation}, where the modality is present but corrupted ($M_{m,i}=1$ with unreliable $E_{m,i}$). The resulting multimodal decision rule is
\[
p(y_i \mid \{E_{m,i},M_{m,i},Q_{m,i}\}_{m\in\mathcal{M}}).
\]

\noindent\textbf{Alignment-breaking control.}
Let \textbf{\CleanQ{}} denote the observed quality values used at inference. 
\textbf{\BrokenQ{}} is formed by shuffling $Q_{m,i}$ across held-out test rows where modality $m$ is present, while keeping evidence $E$ and availability $M$ fixed. This preserves the marginal distribution of quality values and the availability pattern, but breaks the instance-wise link between a quality estimate and its corresponding evidence. For example, a quality score from clean audio may be assigned to a different held-out utterance; a performance drop then indicates that the fused prediction relied on the alignment between audio quality and audio evidence.

To isolate quality effects from missingness, evaluation is restricted to fully observed instances,
\[
\mathcal{D}_{\mathrm{FULL}}=\{\, i \mid \forall m\in\mathcal{M},\ M_{m,i}=1 \,\}.
\]
On this set, availability is constant, so any \CleanQ{} versus \BrokenQ{} difference reflects dependence on quality--evidence alignment rather than missing modalities.

\noindent\textbf{Identifiability statistic.}
Let $S(\cdot)$ denote the evaluation metric, taken to be Balanced Accuracy (equivalent to Unweighted Average Recall in the binary case) \cite{brodersen2010balanced,schuller2013paralinguistics}. For each permutation $k$, let $Q^{(k)}$ denote a \BrokenQ{} sample obtained by shuffling $Q_{m,i}$ within
\[
\{\, i \in \mathrm{TEST} : M_{m,i}=1 \,\}
\]
for each modality $m$, while keeping $E$ and $M$ fixed. The permutation gap is
\begin{equation}
\Delta_{\mathrm{perm}}
=
S(\CleanQm)
-
\mathbb{E}_{k}[S(\BrokenQm^{(k)})].
\label{eq:perm_gap}
\end{equation}
If the trained fusion rule uses quality in an instance-specific way, \CleanQ{} should outperform its shuffled controls and $\Delta_{\mathrm{perm}}$ should be positive.

Define $s_0=S(\CleanQm)$ and $s_k=S(\BrokenQm^{(k)})$. We compute a one-sided permutation value using the unbiased estimator of Phipson and Smyth \cite{phipson2010permutation}: $p_{\mathrm{perm}}=(1+\sum_k \mathbf{1}[s_k\ge s_0])/(K+1)$. Under the null that predictions are invariant to quality shuffling, \CleanQ{} and \BrokenQ{} scores are exchangeable.

\begin{figure}[!t]
    \centering
    \includegraphics[width=.96\columnwidth]{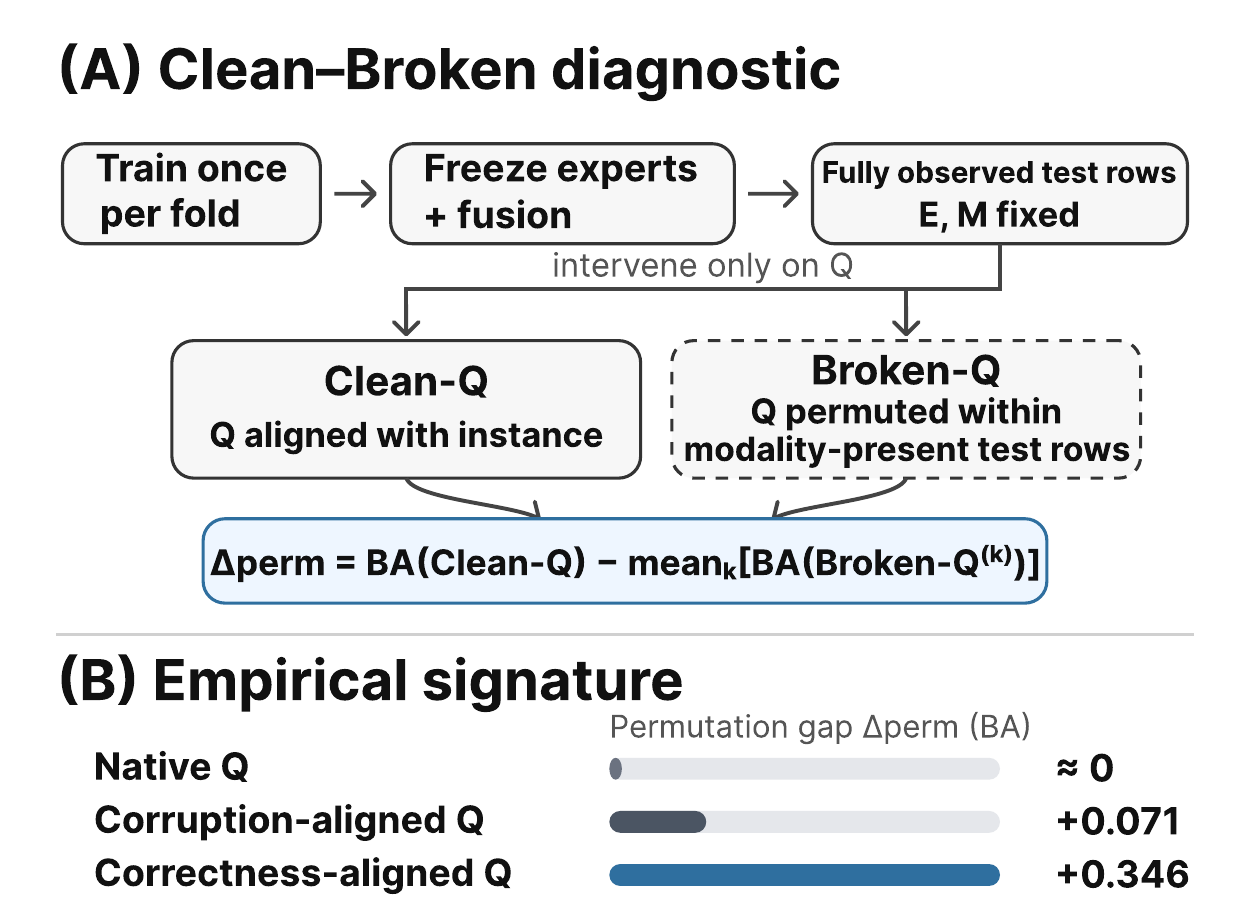}
    \caption{
    Clean--Broken diagnostic and empirical signature. Panel A freezes experts and fusion while changing only quality alignment; evidence $E$ and availability $M$ remain fixed. Panel B shows near-zero native-quality gaps and positive aligned-control gaps.
    }
    \label{fig:pipeline_diagnostic}
\end{figure}

Figure~\ref{fig:pipeline_diagnostic} summarizes the diagnostic in two parts: Panel A shows the test-time intervention on quality while evidence and availability remain fixed, and Panel B previews the empirical signature reported in Section~\ref{sec:results}: native quality produces near-zero gaps, whereas aligned positive controls produce positive gaps.

\noindent\textbf{Interpretation and structural preconditions.}
The null hypothesis is not that quality is useless in general; it is that this trained fusion rule is invariant to quality--instance alignment at inference time. A positive permutation gap indicates that quality influences fused decisions under fixed experts and fusion parameters, while a near-zero gap indicates that shuffling quality does not measurably change the decision rule.

A near-zero gap is most informative when quality-aware routing had room to matter. We therefore report oracle headroom,
\begin{equation}
\text{Headroom}=S(\text{Oracle})-S(\CleanQm),
\label{eq:headroom}
\end{equation}
where \textbf{Oracle} selects the unimodal expert with the highest true-class confidence for each instance. We also report competitiveness, $\Delta_i=p_{(1),i}(y_i)-p_{(2),i}(y_i)$, and quality--correctness alignment, $\rho_m=\operatorname{corr}(Q_{m,i},\mathbf{1}(\hat{y}_{m,i}=y_i))$. Together, these quantities describe whether quality \emph{could} matter; the permutation gap tests whether it \emph{does} matter under the frozen fusion rule.

\section{Datasets}

We evaluate the diagnostic on two complementary multimodal datasets. StressID is the primary setting because it targets stress recognition and contains natural, asymmetric modality availability across speech, facial video, and physiology. CMU-MOSEI serves as a secondary boundary case: it is a different multimodal prediction task with nearly complete language, acoustic, and visual modalities, allowing us to test whether the Clean--Broken pattern persists when missingness is largely absent.

\noindent\textbf{StressID.}
StressID \cite{chaptoukaev2023stressid} contains synchronized facial video, speech, and physiological signals from 65 participants performing 11 tasks. Stress labels are derived from self-assessments of stress/relaxation and valence/arousal. The released dataset includes 711 annotated physiological recordings, 587 annotated videos, and 385 annotated audio recordings, totaling over 39 hours of annotated data. Its key property for our diagnostic is heterogeneous availability: audio is present only for the seven interactive talking tasks, and additional video or audio recordings are missing because of acquisition failures.

\noindent\textbf{CMU-MOSEI.}
CMU-MOSEI \cite{zadeh2018MOSEI} contains 23{,}453 annotated opinion segments from 3{,}228 videos and 1{,}000 speakers across 250 topics. Each segment is annotated for sentiment on a $[-3,3]$ scale and includes aligned language, acoustic, and visual features, including COVAREP acoustic features and FACET visual features. Unlike StressID, CMU-MOSEI is nearly fully observed in our processed setting, so we use it as a boundary case for quality alignment rather than as a realistic modality-outage benchmark.

\section{Experimental protocol}
\label{sec:experimental}

Our protocol is designed to keep three quantities separate throughout evaluation: modality evidence, modality availability, and quality estimates. We first build an aligned instance table for each dataset, then apply group-disjoint splits, train unimodal experts, and finally test whether changing only the quality values affects frozen fusion decisions.

\noindent\textbf{Aligned instance table.}
For each dataset, we construct a fixed-order table with one row per StressID task instance or CMU-MOSEI segment. Each row stores the modality evidence $E$, availability masks $M$, quality values \(Q\), and the label. In StressID, physiology instance IDs define the row order because physiology provides the most consistent coverage across tasks; audio and video embeddings are joined by exact ID match. In CMU-MOSEI, opinion segments aligned to the official label key define the row order. Missing modality feature vectors are zero-filled only for representation consistency and are marked as unavailable with $M_m=0$.

\noindent\textbf{Unimodal representations.}
For StressID, we extract frozen pretrained embeddings from each raw signal stream. Audio embeddings are obtained using Wav2Vec2-base \cite{baevski2020wav2vec}, facial embeddings using a pretrained AffectNet-based encoder \cite{AffectNet}, and physiological embeddings using the MOMENT-1-large time-series encoder \cite{MOMENT}. ECG and EDA are used for physiology; respiration is excluded because raw signals were unavailable. Frame- or window-level features are mean-pooled to produce one vector per instance. Encoders remain frozen, so only the unimodal classifiers and fusion components are trained. For CMU-MOSEI, we use the provided aligned language, acoustic (COVAREP), and visual (FACET) features.

\noindent\textbf{Cross-validation.}
We use $5$ seeds $\times$ $5$ StratifiedGroupKFold \cite{pedregosa2011scikit} splits, yielding 25 folds. Splits are subject-disjoint for StressID and video-disjoint for CMU-MOSEI, preventing identity leakage between training and held-out test rows.

\noindent\textbf{Quality construction.}
Raw signal-derived quality metrics are computed once per modality and aligned to the fixed row order (Table~\ref{tab:rawq_defs}). For each seed, fold, and modality $m$, quantile scaling is fit only on training rows where that modality is observed, i.e., rows with $M_{m,i}=1$. The fitted scaler is then applied unchanged to held-out test rows. Missing or non-finite quality values are mapped to $0$ after scaling. The resulting aligned quality values define \CleanQ{}; \BrokenQ{} is formed by permuting held-out test quality values only among rows where the corresponding modality is observed.

\begin{table}[!t]
\centering
\caption{Raw quality definitions. $M_m$ denotes whether modality $m$ is observed; $Q_m$ denotes the fold-scaled quality value for modality $m$. Missing or non-finite raw quality values are mapped to zero after train-only scaling. Higher $Q_m$ indicates higher estimated reliability.}
\label{tab:rawq_defs}
\footnotesize
\setlength{\tabcolsep}{3pt}
\renewcommand{\arraystretch}{0.98}
\begin{tabular}{
  >{\raggedright\arraybackslash}p{0.14\columnwidth}
  p{0.56\columnwidth}
  >{\raggedright\arraybackslash}p{0.22\columnwidth}
}
\toprule
Modality & RAWQ features ($q_0, q_1$) & Failure mode \\
\midrule

Audio &
\begin{tabular}[t]{@{}l@{}}
\strut $q_0 = 1-\mathrm{clip\_frac}$\\
$q_1 = 10\log_{10}(p_{90}^2/p_{10}^2)$\strut
\end{tabular}
&
Low~SNR; clipping
\\
\addlinespace[0.6ex]

Physio &
\begin{tabular}[t]{@{}l@{}}
\strut $q_0 = \log_{10}(\mathrm{BP}_{0.5\!-\!40}/\mathrm{BP}_{40\!-\!100})$\\
$q_1 = 1-\max(\pi_{\mathrm{ECG}},\pi_{\mathrm{EDA}},\pi_{\mathrm{RR}})$\strut
\end{tabular}
&
Dropout; flatline; spikes
\\
\addlinespace[0.6ex]

Video &
\begin{tabular}[t]{@{}l@{}}
\strut $q_0 = 1-(\tilde{u}+\tilde{o})$\\
$q_1 = \widetilde{\mathrm{Var}}(\nabla^2 I)$\strut
\end{tabular}
&
Under/over-exp.; blur
\\

\bottomrule
\end{tabular}
\end{table}

\noindent\textbf{Training and evaluation.}
For each fold, unimodal experts are trained on observed training rows and produce probabilistic outputs for both training and held-out test rows. Fusion models are trained only after the unimodal experts are fixed. We then evaluate \CleanQ{} and \BrokenQ{} on fully observed held-out test instances, so any difference reflects quality alignment rather than missing modalities. Results are reported as mean $\pm$ standard deviation of Balanced Accuracy across the 25 folds. Code and precomputed artifacts will be released after publication.

\section{Fusion methods}
\label{sec:fusion}

We use decision-level fusion so that the diagnostic tests routing among fixed unimodal experts rather than changes in representation learning. For each seed$\times$fold and modality $m$, a unimodal expert is trained only on training rows where that modality is observed. The expert is then frozen and produces a posterior probability $p_{m,i}=p_m(y_i=1\mid E_{m,i})$ for each instance. We instantiate these experts with two classifiers: logistic regression (LR), a stable linear diagnostic model, and histogram-based gradient boosting (HGB), a nonlinear robustness check. The fusion models receive only the vector of expert posteriors,
\[
\mathbf{p}_i=[p_{m,i}]_{m\in\mathcal{M}},
\]
so any Clean--Broken difference reflects how the fusion rule uses quality information to weight fixed expert predictions.

We evaluate two fusion families: a decision-level late-fusion rule and a conditioning-aware mixture of experts. These instantiate standard late-fusion and reliability-conditioned routing ideas used in multimodal learning \cite{Baltrusaitis.2018,han2021,tsai.2019,soleymani.2022}.

\noindent\textbf{Late fusion.}
The quality-free baseline averages expert probabilities uniformly:
\[
p_{\NoQm}(y_i=1)=\frac{1}{|\mathcal{M}|}\sum_{m\in\mathcal{M}} p_{m,i}.
\]
The quality-aware version instead weights each expert by its normalized quality value,
\[
w_{m,i}=\frac{Q_{m,i}}{\sum_{m'} Q_{m',i}}, 
\quad
p_{\CleanQm}(y_i=1)=\sum_{m\in\mathcal{M}} w_{m,i}\, p_{m,i}.
\]
If all quality values are zero or missing after preprocessing, we fall back to uniform averaging.

\noindent\textbf{Conditioning-aware mixture of experts.}
The mixture-of-experts model learns a linear softmax router that maps availability and quality features to expert weights:
\[
\mathbf{w}_i=\mathrm{softmax}(X_i W + b),
\]
where $X_i=[\,M_{m,i},Q_{m,i}\,]_{m\in\mathcal{M}}$ concatenates availability and quality in a fixed modality order. The final prediction is
\[
p(y_i=1)=\sum_{m\in\mathcal{M}} w_{m,i}\,p_{m,i}.
\]
The router is trained on training rows with aligned quality values and then frozen. At held-out test time, we change only the quality values: \CleanQ{} keeps quality matched to its original instance, whereas \BrokenQ{} shuffles quality across eligible test rows. Evaluation is restricted to fully observed rows, so the Clean--Broken gap measures dependence on quality alignment rather than missing modalities.

\section{Results}
\label{sec:results}

\noindent\textbf{Summary.}
Native quality estimates barely change fused predictions, whereas aligned positive controls do. Thus, quality affects decisions only when it identifies the reliable modality for the current instance.

\subsection{StressID}

\noindent\textbf{Opportunity for quality-aware routing.}
Because we use subject-disjoint folds rather than the original random task splits, absolute Balanced Accuracy is not directly comparable. Table~\ref{tab:uni} shows that audio is strongest on fully observed StressID test rows (LR: $0.592\pm0.08$; HGB: $0.569\pm0.05$). Table~\ref{tab:stressid_structure} shows substantial expert disagreement and competitiveness (median $\Delta\approx0.20$), so routing could affect decisions. However, quality--correctness alignment is near zero ($\rho\approx0$), meaning that native quality estimates do not reliably identify the correct expert.

\begin{table}[t]
\centering
\caption{StressID unimodal Balanced Accuracy (mean$\pm$std) on fully observed test rows. LR = logistic regression; HGB = histogram-based gradient boosting.}
\label{tab:uni}
\footnotesize
\setlength{\tabcolsep}{6pt}
\renewcommand{\arraystretch}{0.95}
\begin{tabular}{lcc}
\toprule
Expert & LR & HGB \\
\midrule
\textbf{Audio} & $\mathbf{0.592}\pm0.08$ & $\mathbf{0.569}\pm0.05$ \\
Physio  & $0.484\pm0.08$          & $0.488\pm0.09$ \\
Video   & $0.454\pm0.09$          & $0.466\pm0.08$ \\
\bottomrule
\end{tabular}
\end{table}
\begin{table}[t]
\centering
\caption{StressID structural diagnostics on fully observed test rows. Near-tie: $\Delta_i<0.05$; Disagr.: different hard predictions; $\rho$: quality--correctness alignment.}
\label{tab:stressid_structure}
\footnotesize
\setlength{\tabcolsep}{4pt}
\begin{tabular*}{\columnwidth}{@{\extracolsep{\fill}}lcc@{}}
\toprule
Metric & LR & HGB \\
\midrule
Median $\Delta$ & $0.209 \pm 0.08$ & $0.204 \pm 0.06$ \\
Near-tie (\%)   & $28.1 \pm 8$     & $15.4 \pm 6$ \\
Disagr. (\%)    & $71.5 \pm 8$     & $67.7 \pm 11$ \\
Align. $\rho$   & $-0.03$          & $-0.02$ \\
\bottomrule
\end{tabular*}
\end{table}

\noindent\textbf{Native quality diagnostic.}
Table~\ref{tab:stressid_ident} shows near-zero Clean--Broken gaps despite room for better per-example routing: logistic regression $-0.002\pm0.06$, histogram-based gradient boosting $-0.011\pm0.06$, and mixture of experts $-0.003\pm0.02$. Shuffling native quality therefore does not measurably change frozen fusion decisions.

\begin{table}[t]
\centering
\caption{StressID Clean--Broken diagnostic ($K=200$) on fully observed test rows. Gaps remain near zero despite oracle headroom.}
\label{tab:stressid_ident}
\footnotesize
\setlength{\tabcolsep}{8pt}
\renewcommand{\arraystretch}{0.95}
\begin{tabular}{@{}lccc@{}}
\toprule
Family & $\bm{\Delta_{\mathrm{perm}}}$ & $p_{\mathrm{med}}$ & Oracle Headroom \\
\midrule
LR  & $-0.002\pm0.06$ & 0.57 & $0.361\pm0.08$ \\
HGB & $-0.011\pm0.06$ & 0.71 & $0.352\pm0.08$ \\
MoE & $\mathbf{-0.003}\pm0.02$ & 0.66 & $0.372\pm0.07$ \\
\bottomrule
\end{tabular}
\end{table}

\noindent\textbf{Stress tests and positive controls.}
Increasing expert competition and degrading audio confidence in low-quality regions still leaves permutation gaps near zero (max $|\Delta_{\mathrm{perm}}|\le0.0024$, median $p\ge0.50$), indicating that native quality remains decision-irrelevant under stronger routing pressure. In contrast, Table~\ref{tab:diagnostic_summary} verifies diagnostic sensitivity: corruption- and correctness-aligned quality produce positive gaps ($+0.071\pm0.03$ and $+0.346\pm0.06$), so frozen fusion rules use quality when it ranks experts correctly.

\begin{table}[t]
\centering
\caption{StressID positive controls. Alignment with corruption or correctness yields significant positive gaps.}
\label{tab:diagnostic_summary}
\footnotesize
\setlength{\tabcolsep}{5pt}
\renewcommand{\arraystretch}{0.95}
\begin{tabular}{lcc}
\toprule
Control & $\Delta_{\mathrm{perm}}$ & $p_{\mathrm{med}}$ \\
\midrule
Corruption ($k{=}0.5$, $Q^{\mathrm{syn}}$) 
& $+0.071\pm0.03$ & 0.020 \\

Sufficiency (aligned $Q^{\mathrm{align}}$) 
& $+0.346\pm0.06$ & 0.005 \\
\bottomrule
\end{tabular}
\end{table}

\subsection{CMU-MOSEI boundary case}

We use CMU-MOSEI as a near-fully observed boundary case for quality alignment, not as realistic outage validation. Language dominates acoustic and visual experts (Balanced Accuracy: $0.711$ vs.\ $0.591/0.590$), with moderate competitiveness (median $\Delta=0.128$) but weak quality--correctness alignment ($\rho=-0.01$). The Clean--Broken gap is small and non-significant ($0.004\pm0.004$, median $p=0.07$) despite oracle headroom ($0.216\pm0.006$), matching StressID: better routing is possible, but native quality does not identify the reliable modality.

\section{Discussion}
\label{sec:discussion}

Native quality estimates do not measurably affect fused decisions in either dataset, despite room for better per-example routing. This is not a fusion-capacity limitation: StressID positive controls show that the same frozen rules change decisions when quality tracks corruption or unimodal correctness. The diagnostic separates three often-conflated questions: whether better per-instance routing is possible, whether the fusion rule can use a routing signal, and whether native quality supplies that signal. Here, the first two hold, but the third largely does not. Thus, quality awareness in architecture does not imply quality dependence in decisions; reliability must predict which expert is correct for the current instance, not merely summarize generic signal cleanliness. 

Dataset structure bounds this interpretation. StressID contains asymmetric missingness, so availability-aware routing is meaningful. CMU-MOSEI is nearly fully observed ($99.97\%$ fully observed rows), making it a quality-alignment boundary case rather than a realistic outage benchmark. It should be read as a cross-task alignment check, not as evidence about natural modality dropout. Extending the diagnostic to early- or mid-fusion attention models would require an intervenable reliability signal or a representation-level intervention.

\noindent\textbf{Limitations.}
The structural diagnostics require labels and are evaluation tools, not deployment mechanisms. Fully observed evaluation isolates quality effects but not quality--outage interactions, and positive controls are StressID-only. An ideal follow-up dataset would combine natural modality dropout, broader within-modality quality variation, and modality-specific positive controls, enabling the same diagnostic to test both outage and degradation.

\section{Conclusion}
\label{sec:conclusion}
Quality-aware multimodal fusion helps only when reliability estimates identify which input stream to trust for the current example. We tested this dependence with a leakage-safe intervention: after training, experts and the fusion rule are frozen, and test-time reliability estimates are shuffled while sensory evidence and modality availability remain fixed. If reliability guides decisions, this shuffle should change predictions or reduce balanced accuracy. Across StressID and a near-fully observed CMU-MOSEI boundary case, shuffling native reliability estimates barely changed fused decisions, even though selecting the best modality for each example showed that better routing was possible. StressID positive controls changed decisions when reliability tracked corruption or unimodal correctness, showing that the same frozen fusion rules can use reliability when it is aligned with expert correctness. Thus, reliability--correctness alignment, not fusion capacity, is the bottleneck; quality-aware fusion claims should be validated with alignment-breaking tests.

\clearpage

\section{Acknowledgments}
We thank the anonymous reviewers and meta-reviewer for their constructive feedback. We also thank the creators and maintainers of the StressID and CMU-MOSEI datasets for making these resources available to the research community.

\section{Generative AI Use Disclosure}
Generative AI tools were used solely to assist with proofreading and improving grammar, clarity, and overall writing flow of the manuscript. These tools were not used to generate experimental results, analyze data, design the methodology, or produce substantive scientific content. All research design decisions, experiments, analyses, and interpretations were conducted by the authors.

\bibliographystyle{IEEEtran}
\bibliography{mybib}

\end{document}